\documentclass[sigconf]{acmart}
\usepackage{amsmath}
\usepackage{breqn}
\usepackage{array}
\newcolumntype{P}[1]{>{\centering\arraybackslash}p{#1}}
\newcolumntype{M}[1]{>{\centering\arraybackslash}m{#1}}
\usepackage{multirow}
\usepackage{paralist}
\usepackage{comment}

\setcopyright{none}
\renewcommand\footnotetextcopyrightpermission[1]{}

\definecolor{dkself}{rgb}{0.6,0.2,0.6}

\AtBeginDocument{%
  \providecommand\BibTeX{{%
    \normalfont B\kern-0.5em{\scshape i\kern-0.25em b}\kern-0.8em\TeX}}}

\setcopyright{acmcopyright}
\copyrightyear{2021}
\acmYear{2021}
\acmDOI{10.1145/xxxxxxx.xxxxxxx}

\begin{document}
\thispagestyle{empty}

%%
%% The "title" command has an optional parameter,
%% allowing the author to define a "short title" to be used in page headers.
%\title{Backdoor Attacks and the Explainability of Triggers for Graph Neural Networks}
\title{Explainability-based Backdoor Attacks \\Against Graph Neural Networks} %\jason

%%
%% The "author" command and its associated commands are used to define
%% the authors and their affiliations.
%% Of note is the shared affiliation of the first two authors, and the
%% "authornote" and "authornotemark" commands
%% used to denote shared contribution to the research.
% \author{Ben Trovato}
% \authornote{Both authors contributed equally to this research.}
% \email{trovato@corporation.com}
% \orcid{1234-5678-9012}
% \author{G.K.M. Tobin}
% \authornotemark[1]
% \email{webmaster@marysville-ohio.com}
% \affiliation{%
%   \institution{Institute for Clarity in Documentation}
%   \streetaddress{P.O. Box 1212}
%   \city{Dublin}
%   \state{Ohio}
%   \country{USA}
%   \postcode{43017-6221}
% }
\author{Jing Xu}
\email{j.xu-8@tudelft.nl}
\affiliation{%
   \institution{Delft University of Technology}
   \country{Netherlands}
}

\author{Minhui (Jason) Xue}
\email{jason.xue@adelaide.edu.au}
\affiliation{%
   \institution{The University of Adelaide}
   \country{Australia}
}

\author{Stjepan Picek}
\email{s.picek@tudelft.nl}
\affiliation{%
   \institution{Delft University of Technology}
   \country{Netherlands}
}

\begin{abstract}
Backdoor attacks represent a serious threat to neural network models. A backdoored model will misclassify the trigger-embedded inputs into an attacker-chosen target label while performing normally on other benign inputs. 
There are already numerous works on backdoor attacks on neural networks, but only a few works consider graph neural networks (GNNs). As such, there is no intensive research on explaining the impact of trigger injecting position on the performance of backdoor attacks on GNNs.

To bridge this gap, we conduct an experimental investigation on the performance of backdoor attacks on GNNs. We apply two powerful GNN explainability approaches to select the optimal trigger injecting position to achieve two attacker objectives -- high attack success rate and low clean accuracy drop. Our empirical results on benchmark datasets and state-of-the-art neural network models demonstrate the proposed method's effectiveness in selecting trigger injecting position for backdoor attacks on GNNs. For instance, on the node classification task, the backdoor attack with trigger injecting position selected by GraphLIME reaches over $84 \%$ attack success rate with less than $2.5 \%$ accuracy drop.
\end{abstract}

%%
%% The code below is generated by the tool at http://dl.acm.org/ccs.cfm.
%% Please copy and paste the code instead of the example below.
%%

\begin{comment}
\begin{CCSXML}
<ccs2012>
   <concept>
       <concept_id>10002978</concept_id>
       <concept_desc>Security and privacy</concept_desc>
       <concept_significance>500</concept_significance>
       </concept>
   <concept>
       <concept_id>10002978.10003022</concept_id>
       <concept_desc>Security and privacy~Software and application security</concept_desc>
       <concept_significance>300</concept_significance>
       </concept>
 </ccs2012>
\end{CCSXML}

\ccsdesc[500]{Security and privacy}
\ccsdesc[300]{Security and privacy~Software and application security}

\end{comment}
%%
%% Keywords. The author(s) should pick words that accurately describe
%% the work being presented. Separate the keywords with commas.

%\keywords{Graph Neural Networks, Backdoor Attacks, Explainability}

%% A "teaser" image appears between the author and affiliation
%% information and the body of the document, and typically spans the
%% page.
% \begin{teaserfigure}
%   \includegraphics[width=\textwidth]{sampleteaser}
%   \caption{Seattle Mariners at Spring Training, 2010.}
%   \Description{Enjoying the baseball game from the third-base
%   seats. Ichiro Suzuki preparing to bat.}
%   \label{fig:teaser}
% \end{teaserfigure}

%%
%% This command processes the author and affiliation and title
%% information and builds the first part of the formatted document.

\maketitle

\thispagestyle{empty}
\section{Introduction}
\label{sec:introduction}

Many real-world data can be modeled as graphs, such as social relations or protein structures. Graph Neural Networks (GNNs) have emerged as state-of-the-art for machine learning on graphs due to their superior ability to incorporate information from neighboring nodes in the graph recursively by capturing both graph structure and node features~\cite{Wu2021}.
However, similar to Convolutional Neural Networks (CNNs), GNNs are also vulnerable to adversarial attacks, one of which is the backdoor attack. Since GNNs are used increasingly more for security applications, it is important to study the backdoor attack on GNNs. Otherwise, there will remain security concerns. For instance, in a Bitcoin transaction ego network~\cite{DBLP:journals/corr/abs-1908-02591}, where the nodes are the transactions, and the edge between two nodes indicates the flow of Bitcoin from one transaction to another, the attacker can attack the GNNs to classify an illegal transaction as a legal one.

In the backdoor attacks on GNNs, the trigger injecting position impacts the attack's performance in terms of the attack success rate and clean accuracy drop. Recently, some works are exploiting the vulnerabilities of GNNs to backdoor attacks with different trigger injecting position selecting strategies~\cite{zhang2020backdoor, xi2021graph}. However, these works either select trigger injecting position randomly, in which situation the attack may be easily detected by the defender~\cite{zhang2020backdoor}, or use a computationally intensive algorithm to get the trigger injecting position, as shown in~\cite{xi2021graph}. If we know how to quickly select the optimal (or close to optimal) trigger injecting position in backdoor attacks on GNNs, we can achieve high attack performance and good evasion of the defender's detection mechanisms. Further, we can develop more robust GNN models. Unfortunately, since graph data have characteristics of complex relationships and interdependencies between objects, common explainability approaches for CNNs, such as Shapley value~\cite{pmlr-v97-ancona19a}, are not suitable to explain the predictions of GNNs to select the optimal trigger injecting position.

%\textbf{Backdoor attacks} - 
Deep Neural Networks (DNNs) are vulnerable to backdoor attacks~\cite{DBLP:conf/ndss/LiuMALZW018,9186317}. Specifically, a backdoored neural network classifier produces attacker-desired behaviors when a trigger is injected into a testing example. Several studies showed that GNNs are also vulnerable to backdoor attacks. Zhang et al. proposed a subgraph-based backdoor attack to GNNs for graph classification task~\cite{zhang2020backdoor}. Xi et al. presented a subgraph-based backdoor attack to GNNs, but this attack can be instantiated for both node classification and graph classification tasks~\cite{xi2021graph}.

%\textbf{Interpretability and Explanations for Neural Networks} - 
Interpretability methods for non-graph neural networks can be grouped into two main categories. Methods in the first group construct simple proxy models of the original neural networks by learning a locally faithful approximation around the prediction~\cite{DBLP:conf/kdd/Ribeiro0G16}. Methods in the second group identify important aspects of the computation~\cite{DBLP:conf/eccv/ZeilerF14, DBLP:conf/icml/SundararajanTY17}. Since these approaches are not suitable for explaining predictions made by GNNs, recently, some works try to interpret GNNs. Ying et al. proposed to utilize mutual information to find a subgraph with associated features for interpreting the predicted label of a node or graph being explained~\cite{DBLP:conf/nips/YingBYZL19}. Huang et al. presented a method utilizing predicted labels from both the node being explained and its neighbors, which enables to capture more local information around the node and give a finite number of features as explanations in an intuitive way~\cite{DBLP:journals/corr/abs-2001-06216}.

In this paper, we propose utilizing powerful neural network approaches that explain predictions made by GNNs to understand the performance of backdoor attacks on GNNs. Indeed, backdoor attacks on GNNs have been presented~\cite{zhang2020backdoor, xi2021graph} but how to quickly select the optimal trigger injecting position and what is the impact of different trigger injecting position on the attack performance have not been explored. In this paper, we seek to bridge this gap.
To the best of our knowledge, this work represents the first study on the explainability of triggers for backdoor attacks on GNNs.
Our contributions can be summarized as follows:
\begin{compactitem}
    \item We utilize GNNExplainer, an approach for explaining predictions made by GNNs, to analyze the impact of trigger injecting position for the backdoor attacks on GNNs for the graph classification task.
    \item We propose a new backdoor attack on GNNs for the node classification task, which uses a subset of node features as a trigger pattern. Additionally, we explore GraphLIME, a local interpretable model explanation for graphs, to explore the proposed backdoor attack on the node classification task through modifying a different subset of node features.
\end{compactitem}
We conduct an empirical study of the explainability of backdoor attack triggers on GNNs using various state-of-the-art GNNs and four benchmark datasets.

% Roadmap - The remainder of the paper proceeds as follows. Section~\ref{sec:background} introduces background concepts. Section~\ref{sec:explainable_backdoor_attack} first presents how to apply the GNNExplainer method to analyze the backdoor attack to GNNs on the graph classification task. Next, we present a new backdoor attack method to GNNs on the node classification task and use the GraphLIME approach to explain the impact of different feature selecting strategies. Section~\ref{sec:experiment} provides experimental results. We conclude the paper in Section~\ref{sec:conclusions} where we also briefly discuss future research directions.
\thispagestyle{empty}
\section{Background}
\label{sec:background}
%In this section, we start with a brief introduction to graph neural networks. Afterward, we discuss backdoor attacks on GNNs. 

\subsection{Preliminaries}
\noindent \textbf{Graph Neural Networks (GNNs).}  GNNs take a graph $G$ as an input, including its structure information and node features, and learn a representation vector (embedding) for each node $v \in G$, $z_v$, or the entire graph, $z_G$. Modern GNNs follow a neighborhood aggregation strategy, where one iteratively updates the representation of a node by aggregating representations of its neighbors. After $k$ iterations of aggregation, a node's representation captures both structure and feature information within its $k$-hop network neighborhood. Formally, the $k$-th layer of a GNN is (e.g., GCN~\cite{kipf2017semi}, GraphSAGE~\cite{DBLP:conf/nips/HamiltonYL17}, and GAT~\cite{velickovic2018graph}):
\begin{equation}
    Z^{(k)} = AGGREGATE(A, Z^{(k-1)};\theta^{(k)}),
    \label{eqn:2.1-1}
\end{equation}
where $Z^{(k)}$ are the node embeddings in the matrix form computed after the $k$-th iteration and the $AGGREGATE$ function depends on the adjacency matrix $A$, the trainable parameters $\theta^{(k)}$, and the previous node embeddings $Z^{(k-1)}$. $Z^{(0)}$ is initialized as $G$'s node features.

For the node classification task, the node representation $Z^{(k)}$ of the final iteration is used for prediction. And for the graph classification task, the READOUT function pools the node embeddings from the final iteration $K$:
\begin{equation}
    z_G = READOUT(Z^{(K)}).
    \label{eqn:2.1-2}
\end{equation}
READOUT can be a simple permutation invariant function such as summation or a more sophisticated graph-level pooling function~\cite{DBLP:conf/nips/YingY0RHL18, DBLP:conf/aaai/ZhangCNC18}.

\noindent \textbf{Graph-Level classification.}  Graph-level classification aims to predict the class label(s) for an entire graph~\cite{DBLP:conf/aaai/ZhangCNC18}. The end-to-end learning for this task can be realized using graph convolutional layers and readout layers. While graph convolutional layers are responsible for extracting high-level node representations, the readout layer collapses node representations of each graph into a graph representation. By applying a multilayer perceptron and a Softmax layer to graph representations, one can build an end-to-end framework for graph classification.

\noindent \textbf{Node-Level classification.}  Given a single graph with partial nodes being labeled and others remaining unlabeled, GNNs can learn a robust model that effectively identifies the class labels for the unlabeled nodes~\cite{kipf2017semi}. In a node-level classification task, there are two types of training settings - inductive and transductive. In an inductive setting, the unlabeled nodes are not seen during training, while in a transductive setting, the test nodes (but not their labels) are also observed during the training process. In this paper, we focus on the transductive node-level classification task.

\noindent \textbf{Explainability tools for GNNs.}  GNNExplainer is a model-agnostic approach for providing explanations on predictions of any GNN-based model. Given a trained GNN model and its prediction(s), GNNExplainer returns an explanation in the form of a small subgraph of the input graph together with a small subset of node features that are most influential for the
prediction(s)~\cite{DBLP:conf/nips/YingBYZL19}. GraphLIME is a local interpretable model explainability method for graphs. More specifically, to explain a node, GraphLIME generates a nonlinear interpretable model from its $N$-hop neighborhood and then computes the most $n$ representative features as the explanations of its prediction using HSIC Lasso~\cite{DBLP:journals/corr/abs-2001-06216}. 

\subsection{Threat Model}
We assume our threat model similar to the existing backdoor attacks, see, e.g.,~\cite{DBLP:conf/ndss/LiuMALZW018}. Given a pre-trained GNN model $\Phi_o$, the adversary forges a backdoored GNN $\Phi$ by perturbing its model parameters without modifying the neural network architecture. We assume the attacker has access to a dataset $D$ sampled from the training dataset. Specifically, in the graph classification dataset, the attacker can inject a trigger (graph) to each intended poisoned training graph and change the label to an attacker-chosen target label. In the node classification dataset, the attacker can inject a feature trigger (feature vector) to each intended poisoned training node and relabel the label to the target label.
Consequently, our attack is a gray-box attack that does not modify the GNN's model architecture but perturbs the model parameters. This represents a realistic model occurring in real-world settings. For instance, if the training dataset is collected from public users, the malicious users under an attacker's control can provide trigger-embedded training data. Note that even without direct access to such data, it is often possible for the attacker to synthesize data to implement backdoor attacks~\cite{DBLP:conf/ndss/LiuMALZW018}.

\section{Explainable Backdoor Attacks}
\label{sec:explainable_backdoor_attack}
%Next, we describe the framework of utilizing GNNs explanation approaches to explain the performance of backdoor attacks to GNNs on two graph tasks - graph classification and node classification. 

\subsection{Backdoor Attacks on Graph Classification}
\begin{figure*}
	\centering
		\includegraphics[scale=.29, page=1]{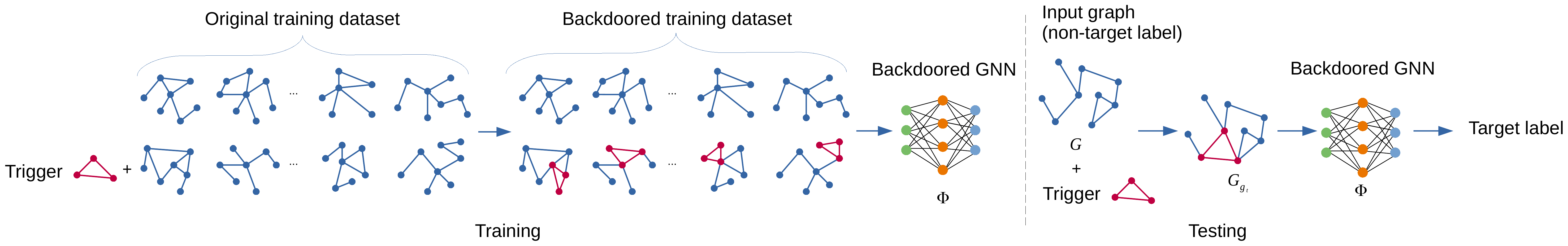}
	\caption{Illustration of subgraph based backdoor attack on GNNs for graph classification task.}
	\label{FIG:1}
\end{figure*}

%Given a pre-trained GNN model $\Phi_o$, the adversary aims to forge a backdoored model $\Phi$ so that $\Phi$ forces to misclassify all the trigger-embedded graphs to the target class $y_t$ while performing normally on other graphs. 
Since most graph classification tasks are implemented by utilizing GNNs to learn the network structure, we focus on subgraph-based backdoor attacks on the graph classification task. Figure~\ref{FIG:1} illustrates the pipeline of subgraph-based backdoor attack on GNNs for the graph classification task. Formally, in the training phase, the attacker injects a trigger (graph) $g_t$ to a subset of the original training dataset and changes their labels to the attacker-chosen target label to obtain the backdoored training dataset. A GNN model trained using the backdoored training dataset is called backdoored GNN $\Phi$. Then in the testing phase, the adversary injects the same trigger to a given graph $G$. If we define such trigger-embedded graph as $G_{g_t}$, the adversary's objective can be defined as:
\begin{equation}
    \left\{\begin{matrix}
    \Phi(G_{g_t}) = y_t\\
    \Phi(G) = \Phi_o(G)
    \end{matrix}\right.
    \label{eqn:3.1-1}
\end{equation}

The first objective in Eq.~\eqref{eqn:3.1-1} means that all the trigger-embedded graphs are required to be misclassified to the target class $y_t$, i.e., attack effectiveness. In contrast, the second objective ensures that the backdoored GNN performs indistinguishably on normal graphs compared to the original GNN, i.e., attack evasiveness. 

It is challenging to find the optimal trigger injecting position so that the adversary can reach several goals: 1) high attack success rate, 2) high accuracy in normal graphs, and 3) difficult to detect by the defender. More precisely, 
\begin{compactitem}
    \item if we sample $t$ nodes from the graph uniformly at random as the trigger nodes in the trigger graph, the trigger will likely be injected into a subgraph that is important for the GNN's final prediction. As a result, the defender can detect the trigger-embedded graphs easily;
    \item to achieve the second objective, we can select a subgraph similar to $g_t$ in the graph as the trigger injecting position. However, the computation of the similarity between graphs is a complex task as subgraph isomorphism is known to be NP-complete. The graph matching algorithms require an exponential time for computation~\cite{DBLP:journals/pami/CordellaFSV04}.
\end{compactitem}

To overcome the above challenges, we utilize GNNExplainer to optimize the trigger injecting position to ensure the attack effectiveness and attack evasiveness at the same time.
\begin{compactitem}
    \item We first apply GNNExplainer to analyze the prediction of GNNs to understand the impact of each structure in the graph on the classification result from GNNs.
    \item Instead of selecting $t$ nodes from the graph uniformly at random as the trigger nodes, we select the $t$ least important nodes in the graph as the trigger injecting position, which results in difficult-to-detect trigger-embedded graphs.
\end{compactitem}

\begin{figure*}
	\centering
		\includegraphics[scale=.33, page=3]{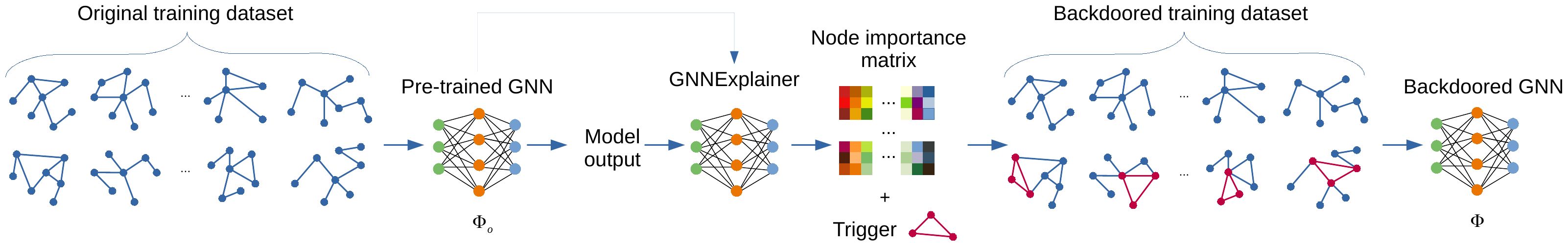}
	\caption{The framework of the backdoor attack on graph classification task based on GNNExplainer.}
	\label{FIG:2}
\end{figure*}

The overall framework of backdoor attack on graph classification task based on GNNExplainer is shown in Figure~\ref{FIG:2}. Given a pre-trained GNN and its predictions, through GNNExplainer, the importance value of nodes for each graph can be computed. Based on the node importance matrix, we select the optimal trigger injecting position for each intended poisoned graph and then train the backdoored GNN.

\thispagestyle{empty}

\subsection{Backdoor Attacks on Node Classification}
 %For a transductive setting, given a graph $G$ and a set of labeled nodes, the GNN's goal is to infer the classes of the remaining unlabeled nodes $V_U$. 
 The currently proposed backdoor attack on GNNs for the node classification task defines triggers as specific subgraphs, i.e., given an arbitrary subgraph $g$ in $G$, by replacing $g$ with the trigger (graph) $g_t$, the adversary attempts to force the unlabeled nodes within $K$ hops to $g$ to be misclassified into the target label $y_t$.
 
 Here, we propose a new method to implement backdoor attacks on GNNs for the node classification task. We assume that the adversary has access to $G$, including the graph structure information $A$ and the node feature information $X$. Each node $v$ in the graph $G$ has its feature vector $x$. Given an arbitrary node in the graph, by changing the value of a subset of its features as a feature trigger, the attacker aims to force the node to be classified to the target class $y_t$ and simultaneously perform normally in other unmodified nodes.
 Formally, the adversary's objective can be defined as:
 \begin{equation}
    \left\{\begin{matrix}
    \Phi(\emph{v}, x_t; G) = y_t\\
    \Phi(\emph{v}, x; G) = \Phi_o(\emph{v}, x; G)
    \end{matrix}\right.
    \label{eqn:3.2-1}
\end{equation}
Here, $x_t$ represents the feature vector with trigger, obtained by changing the features' values in specific dimensions. 

\begin{figure*}
	\centering
		\includegraphics[scale=.35, page=4]{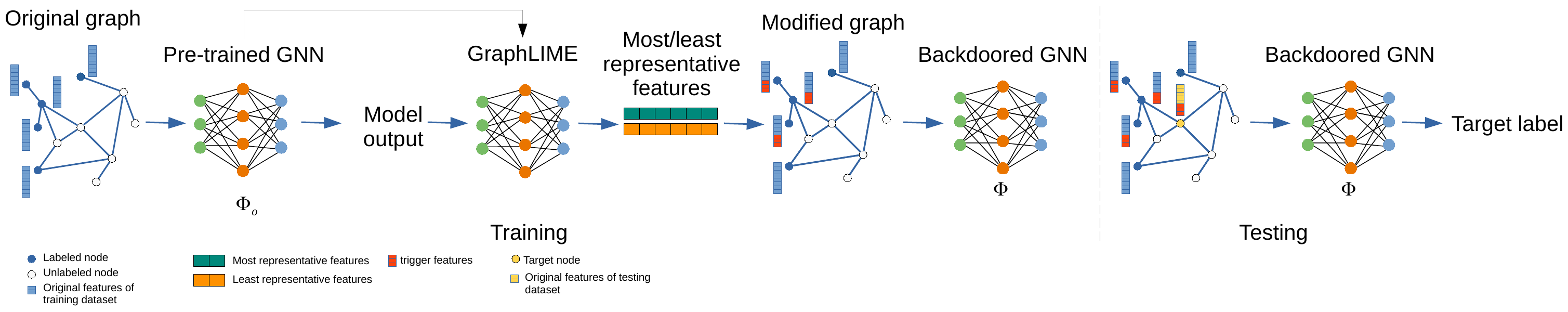}
	\caption{The framework of backdoor attack on node classification task based on GraphLIME.}
	\label{FIG:3}
\end{figure*}

Similar to the graph classification task, these two objectives ensure the attack effectiveness and attack evasiveness. The key point is how to select specific dimensions of a feature vector as a trigger injecting position. Intuitively, we can select $n$ features from the total features uniformly at random and change their values to a fixed value as the feature trigger. We can also use a GNN explainability method - GraphLIME to select the specific feature dimensions:
\begin{compactitem}
    \item We first apply GraphLIME to analyze the output of GNNs on the node classification task to compute the $n$ most/least representative features.
    \item We change the value of the $n$ most/least representative features to a fixed value as the feature trigger and retrain the GNN to get the backdoored GNN.
\end{compactitem}
The overall framework of the proposed backdoor attack on node classification task based on GraphLIME is shown in Figure~\ref{FIG:3}.

\section{Experimental Analysis}
\label{sec:experiment}
We conduct an empirical study of how to utilize the explainability approaches to implement backdoor attacks on state-of-the-art GNN models and benchmark datasets.
%explaining backdoor attack on benchmark datasets, state-of-the-art GNN models, and two common graph tasks - graph classification and node classification.

%\vspace{-3mm}
\subsection{Experimental Setting}

Our experiments were run on an Intel Core i7-8650U CPU processor with $1.90$ GHz frequency and $15.5$ GiB memory. For all the experiments, we use the PyTorch framework.
 
\noindent \textbf{Dataset.}  For the graph classification task, we use two publicly available real-world graph datasets. (i) Mutagenicity~\cite{Morris2020} - molecular structure graphs of mutagen and nonmutagen; (ii) facebook\_ct1~\cite{Morris2020} - a subset of the activity of the New Orleans Facebook community over three months, used to implement the classification task of distinguishing temporal graphs with vertex labels corresponding to observations of a dissemination process and temporal graphs in which the labeling is not a result of a dissemination process. We also use two real-world datasets for node classification task: Cora~\cite{DBLP:journals/aim/SenNBGGE08} and CiteSeer~\cite{DBLP:journals/aim/SenNBGGE08}. Table~\ref{Table:1} shows the statistics of these datasets.

\begin{table*}
\small
 \centering
 \caption{Dataset statistics.}
\begin{tabular}{M{50pt}|M{50pt}|M{50pt}|M{50pt}|M{50pt}|M{110pt}|M{50pt}} 
 \hline
 Datasets & \# Graphs & Avg. \# nodes & Avg. \# edges & Classes & Graphs [Class] & Target class\\
 \hline
 \hline
 Mutagenicity & $4,337$ & $30.32$ & $30.77$ & $2$ & $2,401[0], 1,936[1]$ & $1$ \\
 \hline
 facebook\_ct1 & $995$ & $95.72$ & $269.01$ & $2$ & $498[0], 497[1]$ & $0$ \\
 \hline
 \hline
 \multirow{2}{*}{Cora} & \multirow{2}{*}{$1$} & \multirow{2}{*}{$2,708$} & \multirow{2}{*}{$5,429$} & \multirow{2}{*}{$7$} & $351[0], 217[1], 418[2], 818[3],$ & \multirow{2}{*}{$6$} \\
 & & & & & $426[4], 298[5], 180[6]$ & \\
 \hline
 \multirow{2}{*}{CiteSeer} & \multirow{2}{*}{$1$} & \multirow{2}{*}{$3,327$} & \multirow{2}{*}{$4,608$} & \multirow{2}{*}{$6$} & $264[0], 590[1], 668[2],$ & \multirow{2}{*}{$5$} \\
 & & & & & $701[3], 596[4], 508[5]$ & \\
 \hline
\end{tabular}
\label{Table:1}
\end{table*}

\noindent \textbf{Dataset splits and parameter setting.}  For each graph classification dataset, we sample 2/3 of the graphs as the original training dataset and treat the remaining graphs as the original testing dataset. Among the original training dataset, we randomly sample $\eta$ fraction of graphs to inject the trigger and relabel them with the target label, called the backdoored training dataset. We also inject our trigger to each original testing graph whose label is not the target label to generate the backdoored testing dataset, which is used to evaluate the attack effectiveness. There are several parameters in the attack's implementation: trigger size $s$, trigger density $\rho$, and poisoning intensity $\eta$. We set the trigger size $s$ to be the $\gamma$ fraction of the graph dataset's average number of nodes. Since the trigger size affects the attack effectiveness dramatically, we explore the impact of trigger size on the attack results. At the same time, we set other parameters as: $\rho = 0.8$ and $\eta = 5\%$ following the parameter setting in~\cite{zhang2020backdoor}. We use Erdős-Rényi (ER) model~\cite{Gilbert1959} to generate the trigger with graph density $\rho = 0.8$. 

In the node classification task, we split $20\%$ of the total nodes as the original training dataset (labeled) for each dataset. To generate the backdoored training dataset, we sample $15 \%$ of the original training dataset to inject the feature trigger and relabel these nodes with the target label. The trigger size is set to $10\%$ of the total number of node feature dimensions. We set these parameters as they provided the best results after conducting a tuning phase. %Furthermore, the attack is more likely to evade the detection from a defender with such values.\todo{why? could we delete this sentence?} 
In the node classification task, each node feature has a value of 0 or 1, and here we set the value of the modified node features to 1 (note, the values could also be set to 0).

\noindent \textbf{Models.}  In our experiment, we use the popular GIN~\cite{DBLP:conf/iclr/XuHLJ19} and GraphSAGE~\cite{DBLP:conf/nips/HamiltonYL17} models for the graph classification task as these two methods are the state-of-the-art GNN models. For node classification, we use GAT~\cite{velickovic2018graph} model as the pre-trained GNN model. 

\noindent \textbf{Attack evaluation metrics.}  We use the \textit{attack success rate (ASR)} to evaluate the attack effectiveness. Specifically, in the graph classification task, the ASR measures the proportion of trigger-embedded inputs (the original label is not the target label) that are misclassified by the backdoored GNN into the target class $y_t$ chosen by the adversary. The trigger-embedded inputs are \[D_{g_t} = \left \{ (G_{g_{t1}}, y_1), (G_{g_{t2}}, y_2),  \ldots, (G_{g_{tn}}, y_n) \right \}\] and \[D_{v_t} = \left \{ (v_{x_{t1}}, y_1), (v_{x_{t2}}, y_2), \ldots, (v_{x_{tn}}, y_n) \right \}\] for graph classification and node classification task, respectively. Formally, the ASR can be defined as:
\begin{align*}
 Attack \: Success \: Rate &= \frac{\sum_{i=1}^{n} \mathbb{I}(\Phi(G_{g_{ti}}) = y_t)}{n} \\ 
 &or = \frac{\sum_{i=1}^{n} \mathbb{I}(\Phi(v_{x_{ti}}) = y_t)}{n},
 \label{eqn:4.1-1}
\end{align*}
where $\mathbb{I}$ is an indicator function.

To evaluate the attack evasiveness, we use \textit{clean testing dataset accuracy drop (CAD)}, which is the classification accuracy difference of the original GNN $\Phi_o$ and the backdoored GNN $\Phi$ over the clean testing dataset. 
%Recall that one of the adversary's objectives is that the backdoored GNN performs indistinguishably to the original GNN on normal graphs which are not injected with the trigger, i.e., the \textit{clean testing dataset accuracy drop} should be small.

\thispagestyle{empty}

\subsection{Results for Graph Classification}

This set of experiments evaluate the backdoor attack on GNNs for the graph classification task with the explainability results obtained from GNNExplainer. Based on the node importance matrix from GNNExplainer, we conduct three attacks with different trigger nodes selecting strategies, two among which are proposed here as new strategies: 1) \textbf{Random selecting attack (RSA)} - As in~\cite{zhang2020backdoor}, in this strategy, we sample $t$ nodes from the graph uniformly at random and replace their connection with that in the trigger graph. 2) \textbf{Most important nodes selecting attack (MIA)} - We choose the $t$ most important nodes based on the node importance matrix and replace their connection as that of the trigger graph. 3) \textbf{Least important nodes selecting attack (LIA)} - Instead of selecting the most important $t$ nodes, we select the least important nodes as the trigger nodes.

Table~\ref{Table:2} presents the experimental results of the backdoor attack on the graph classification task based on three attacks. For each result in the table, the first value is ASR, and the second value is CAD.
The results are conducted for the trigger size $\gamma = 0.2$. Finally, we include the performance of two different GNN models - GIN and GraphSAGE. 
We can observe that overall, all three backdoor attacks on GIN achieve high attack effectiveness (each with an attack success rate over $93 \%$) and low clean accuracy drop (each with accuracy drop below $4\%$), while performances in GraphSAGE degrade with attack success rate up to $82 \%$. This may be explained by GIN having a more powerful graph representation capability so the trigger graph can be learned better. The rank between these three attacks, except for the result of the facebook\_ct1 dataset on GraphSAGE, is $LIA \approx RSA > MIA$ in terms of attack effectiveness. The ASR of MIA is lower than the other two attacks probably because after replacing the most important subgraph with the trigger graph, it is more difficult for the GNN to distinguish the graphs of the target class and non-target class. More precisely, GNN needs to dedicate more network capacity to learn specific patterns for each class sample, which negatively influences recognizing the trigger patterns. The result that ASR of RSA and LIA is close means the attacker can inject the trigger to the least important structure of the graph to achieve its goal of being less likely to be detected by the defender.

\begin{figure*}
	\centering
		\includegraphics[scale=.35, page=5]{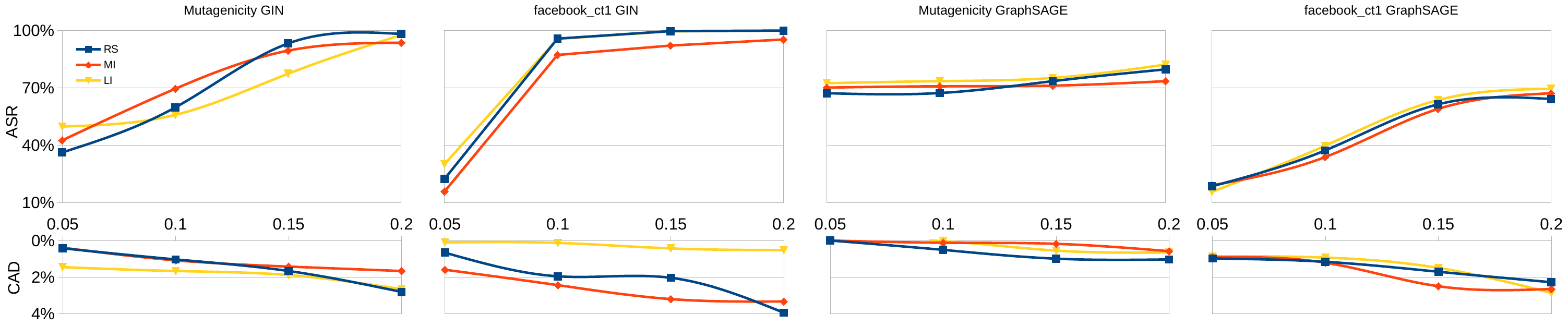}
	\caption{Impact of trigger size $\gamma$ on the attack success rate (ASR) and clean accuracy drop (CAD) of backdoor attack on the graph classification task.}
	\label{FIG:4}
\end{figure*}

We also evaluate the impact of trigger size - $\gamma$ fraction of the average number of nodes on the backdoor attack's performance. Figure~\ref{FIG:4} shows the attack performance under different trigger sizes varying from $5 \%$ to $20 \%$. Obviously, the attack effectiveness of all attacks monotonically increases with the trigger size. This can be easily explained as with larger triggers, backdoored GNNs can better learn the difference between trigger-embedded and normal graphs. Additionally, the clean accuracy drop of all strategies slightly increases as well when the trigger size grows. This may be explained as the trigger size increases, more graph structure information has been modified, and the border between samples from different classes becomes less distinctive, so the performance of the backdoored GNNs on clean dataset drops.

This set of experiments takes on average $13.49$min and $16.72$min to implement backdoor attacks on the GIN model on Mutagenicity and facebook\_ct1 dataset, respectively. The GraphSAGE model takes around $13.37$min and $16.35$min on Mutagenicity and facebook\_ct1 dataset, respectively. Clearly, the process of selecting the optimal trigger injecting position takes a short time on both GNN models on two datasets, e.g., $0.65$s per graph on the GIN model on the Mutagenicity dataset. Consequently, utilizing the GNNExplainer method to select the optimal trigger injecting position for a backdoor attack on GNNs for graph classification task is practical and feasible. What is more, we can select the least important structure of the graph to evade the detection from a defender.

\begin{table*}
\small
 \centering
 \caption{Backdoor attack results on graph classification task based on different trigger nodes selecting strategies.}
\begin{tabular}{M{100pt}|M{50pt}|M{50pt}|M{50pt}|M{50pt}|M{50pt}|M{50pt}} 
 \hline
 \multirow{2}{*}{ASR(\%) | CAD(\%)} & \multicolumn{3}{c|}{GIN} & \multicolumn{3}{c}{GraphSAGE} \\
 \cline{2-7}
 & RSA & MIA & LIA & RSA & MIA & LIA \\
 \hline
 Mutagenicity & $98.24 \;\ 2.80$ & $93.66 \;\ 1.66$ & $97.69 \;\ 2.65$ & $79.73 \;\ 1.03$ & $73.55 \;\ 0.58$ & $82.24 \;\ 0.65$ \\
 \hline
 facebook ct1 & $100 \;\ 3.93$ & $95.35 \;\ 3.32$ & $100 \;\ 0.52$ & $64.23 \;\ 2.27$ & $67.22 \;\ 2.64$ & $69.57 \;\ 2.85$ \\
 \hline
\end{tabular}
\label{Table:2}
\end{table*}

\thispagestyle{empty}

\subsection{Results for Node Classification}
Next, we evaluate the backdoor attack on GNNs for the node classification task under the explainability results of GraphLIME. 
%Specifically, given a subset nodes $V_L$ of a graph $G$, by changing values of subset features $\left \{ x_{v1}, x_{v2},\ldots, x_{vn} \right \}$ for each node $v$, the adversary aims to force the unlabeled nodes with the same feature trigger to be classified to the target label $y_t$. 
Based on the feature importance results of GraphLIME, we propose three attacks with different trigger feature dimension selecting strategies: 1) \textbf{Random selecting attack (RSA)} - select $n$ features from the node feature vector uniformly at random. 2) \textbf{Most important features selecting attack (MIA)} - select the most $n$ representative features and change their values. 3) \textbf{Least important features selecting attack (LIA)} - select the least $n$ representative features and change their values. 

Table~\ref{Table:3} summarizes the attack performance under different trigger feature dimension selecting strategies (the first value is the ASR, and the second value is the CAD). Observe that these three backdoor attacks on GAT obtain high attack success rate, i.e., over $84\%$ and $95\%$ for Cora and CiteSeer, respectively, and low clean accuracy drop at the same time. This is similar to the graph classification task results: the ASR of LIA is close to RSA, while MIA has slightly degraded performance compared to the other two attacks. Therefore, the attacker can select the least representative features of a node to inject the feature trigger to achieve a high attack success rate, and the trigger-embedded node has a lower probability of being detected by the defender.
\begin{table}
\small
 \centering
 \caption{Backdoor attack results on node classification task based on different trigger features selecting strategies.}
\begin{tabular}{M{60pt}|M{50pt}|M{50pt}|M{50pt}} 
 \hline
 \multirow{2}{*}{ASR(\%) | CAD(\%)} & \multicolumn{3}{c}{GAT}\\
 \cline{2-4}
 & RSA & MIA & LIA\\
 \hline
 Cora & $86.01 \;\ 2.23$ & $84.11 \;\ 0.66$ & $84.22 \;\ 1.95$\\
 \hline
 CiteSeer & $96.35 \;\ 1.72$ & $95.28 \;\ 1.39$ & $96.26 \;\ 1.72$\\
 \hline
\end{tabular}
\label{Table:3}
\end{table}

The running time for backdoor attack implementation on the GAT model on two-node classification datasets is on average $27.30$s and $59.02$s, respectively. In the process of selecting the optimal trigger injecting position, it only takes $0.06$s and $0.09$s per node for Cora and CiteSeer dataset, respectively. Similar to the conclusion of graph classification experiments, it is feasible to apply the explainability approach - GraphLIME to select trigger injecting position for a backdoor attack on the node classification task. We emphasize that we do not compare our attack results with the state-of-the-art~\cite{zhang2020backdoor, xi2021graph} because the implementation code for those works is not publicly available.

% \todo{we need to explicitly say why we do not compare with the state-of-the-art?, 1) not publicly available, 2) meet some problems while reproducing the experimental results so not good to compare with the results published in the paper}
% \todo{what are the benefits of using our approach?, 1) mention running time in the results part -> quick to select the trigger position. 2) add running platform informaiton}
% \todo{where is the text about our new backdoor attack?}

\section{Conclusion and Future Work}
\label{sec:conclusions}
This work represents a first step toward the explainability of the impact of trigger injecting position on the performance of backdoor attacks on GNNs. We conduct research on two graph tasks - graph classification and node classification. For the graph classification task, we apply GNNExplainer to select the optimal subgraph in a graph to be replaced by the trigger graph. For the node classification task, we propose a new backdoor attack using a subset of node features as a trigger pattern, and then we apply GraphLIME to choose the optimal subset of node features to change their values to a fixed value as the feature trigger. Through empirical evaluation using benchmark datasets and state-of-the-art models, we verify that our approach can quickly select the optimal trigger injecting position to implement a powerful backdoor attack on GNNs. Furthermore, we see that the attacker can select the least important parts of the graph to inject the trigger, thus reducing the chances of easy detection by the defender. Interesting future work includes: 1) exploring defenses against the backdoor attacks by using the explainability approach, and 2) designing ``clean label'' backdoor poisoning attacks against GNNs where the attacker does not control the sample labeling process.

%%
%% The acknowledgments section is defined using the "acks" environment 
%% (and NOT an unnumbered section). This ensures the proper
%% identification of the section in the article metadata, and the
%% consistent spelling of the heading.
% \begin{acks}
% To Robert, for the bagels and explaining CMYK and color spaces.
% \end{acks}

%%
%% The next two lines define the bibliography style to be used, and
%% the bibliography file.
%\balance

%\bibliographystyle{ACM-Reference-Format}
\thispagestyle{empty}
\bibliographystyle{abbrv}
\bibliography{jingxu}

\begin{thebibliography}{10}

\bibitem{pmlr-v97-ancona19a}
M.~Ancona, C.~Oztireli, and M.~Gross.
\newblock Explaining deep neural networks with a polynomial time algorithm for
  shapley value approximation.
\newblock In K.~Chaudhuri and R.~Salakhutdinov, editors, {\em Proceedings of
  the 36th International Conference on Machine Learning}, volume~97 of {\em
  Proceedings of Machine Learning Research}, pages 272--281. PMLR, 09--15 Jun
  2019.

\bibitem{DBLP:journals/pami/CordellaFSV04}
L.~P. Cordella, P.~Foggia, C.~Sansone, and M.~Vento.
\newblock A (sub)graph isomorphism algorithm for matching large graphs.
\newblock {\em {IEEE} Trans. Pattern Anal. Mach. Intell.}, 26(10):1367--1372,
  2004.

\bibitem{Gilbert1959}
E.~N. Gilbert.
\newblock Random graphs.
\newblock {\em The Annals of Mathematical Statistics}, 30(4):1141--1144, Dec.
  1959.

\bibitem{DBLP:conf/nips/HamiltonYL17}
W.~L. Hamilton, Z.~Ying, and J.~Leskovec.
\newblock Inductive representation learning on large graphs.
\newblock In {\em Advances in Neural Information Processing Systems 30: Annual
  Conference on Neural Information Processing Systems 2017, December 4-9, 2017,
  Long Beach, CA, {USA}}, pages 1024--1034, 2017.

\bibitem{DBLP:journals/corr/abs-2001-06216}
Q.~Huang, M.~Yamada, Y.~Tian, D.~Singh, D.~Yin, and Y.~Chang.
\newblock Graphlime: Local interpretable model explanations for graph neural
  networks.
\newblock {\em CoRR}, abs/2001.06216, 2020.

\bibitem{kipf2017semi}
T.~N. Kipf and M.~Welling.
\newblock Semi-supervised classification with graph convolutional networks.
\newblock In {\em International Conference on Learning Representations (ICLR)},
  2017.

\bibitem{9186317}
S.~{Li}, M.~{Xue}, B.~{Zhao}, H.~{Zhu}, and X.~{Zhang}.
\newblock Invisible backdoor attacks on deep neural networks via steganography
  and regularization.
\newblock {\em IEEE Transactions on Dependable and Secure Computing}, pages
  1--1, 2020.

\bibitem{DBLP:conf/ndss/LiuMALZW018}
Y.~Liu, S.~Ma, Y.~Aafer, W.~Lee, J.~Zhai, W.~Wang, and X.~Zhang.
\newblock Trojaning attack on neural networks.
\newblock In {\em 25th Annual Network and Distributed System Security
  Symposium, {NDSS} 2018, San Diego, California, USA, February 18-21, 2018}.
  The Internet Society, 2018.

\bibitem{Morris2020}
C.~Morris, N.~M. Kriege, F.~Bause, K.~Kersting, P.~Mutzel, and M.~Neumann.
\newblock Tudataset: A collection of benchmark datasets for learning with
  graphs.
\newblock In {\em ICML 2020 Workshop on Graph Representation Learning and
  Beyond (GRL+ 2020)}, 2020.

\bibitem{DBLP:conf/kdd/Ribeiro0G16}
M.~T. Ribeiro, S.~Singh, and C.~Guestrin.
\newblock "why should {I} trust you?": Explaining the predictions of any
  classifier.
\newblock In B.~Krishnapuram, M.~Shah, A.~J. Smola, C.~C. Aggarwal, D.~Shen,
  and R.~Rastogi, editors, {\em Proceedings of the 22nd {ACM} {SIGKDD}
  International Conference on Knowledge Discovery and Data Mining, San
  Francisco, CA, USA, August 13-17, 2016}, pages 1135--1144. {ACM}, 2016.

\bibitem{DBLP:journals/aim/SenNBGGE08}
P.~Sen, G.~Namata, M.~Bilgic, L.~Getoor, B.~Gallagher, and T.~Eliassi{-}Rad.
\newblock Collective classification in network data.
\newblock {\em {AI} Mag.}, 29(3):93--106, 2008.

\bibitem{DBLP:conf/icml/SundararajanTY17}
M.~Sundararajan, A.~Taly, and Q.~Yan.
\newblock Axiomatic attribution for deep networks.
\newblock In D.~Precup and Y.~W. Teh, editors, {\em Proceedings of the 34th
  International Conference on Machine Learning, {ICML} 2017, Sydney, NSW,
  Australia, 6-11 August 2017}, volume~70 of {\em Proceedings of Machine
  Learning Research}, pages 3319--3328. {PMLR}, 2017.

\bibitem{velickovic2018graph}
P.~Veli{\v{c}}kovi{\'{c}}, G.~Cucurull, A.~Casanova, A.~Romero, P.~Li{\`{o}},
  and Y.~Bengio.
\newblock {Graph Attention Networks}.
\newblock {\em International Conference on Learning Representations}, 2018.
\newblock accepted as poster.

\bibitem{DBLP:journals/corr/abs-1908-02591}
M.~Weber, G.~Domeniconi, J.~Chen, D.~K.~I. Weidele, C.~Bellei, T.~Robinson, and
  C.~E. Leiserson.
\newblock Anti-money laundering in bitcoin: Experimenting with graph
  convolutional networks for financial forensics.
\newblock {\em CoRR}, abs/1908.02591, 2019.

\bibitem{Wu2021}
Z.~Wu, S.~Pan, F.~Chen, G.~Long, C.~Zhang, and P.~S. Yu.
\newblock A comprehensive survey on graph neural networks.
\newblock {\em {IEEE} Transactions on Neural Networks and Learning Systems},
  32(1):4--24, Jan. 2021.

\bibitem{xi2021graph}
Z.~Xi, R.~Pang, S.~Ji, and T.~Wang.
\newblock Graph backdoor, 2021.

\bibitem{DBLP:conf/iclr/XuHLJ19}
K.~Xu, W.~Hu, J.~Leskovec, and S.~Jegelka.
\newblock How powerful are graph neural networks?
\newblock In {\em 7th International Conference on Learning Representations,
  {ICLR} 2019, New Orleans, LA, USA, May 6-9, 2019}. OpenReview.net, 2019.

\bibitem{DBLP:conf/nips/YingBYZL19}
Z.~Ying, D.~Bourgeois, J.~You, M.~Zitnik, and J.~Leskovec.
\newblock Gnnexplainer: Generating explanations for graph neural networks.
\newblock In {\em Advances in Neural Information Processing Systems 32: Annual
  Conference on Neural Information Processing Systems 2019, NeurIPS 2019,
  December 8-14, 2019, Vancouver, BC, Canada}, pages 9240--9251, 2019.

\bibitem{DBLP:conf/nips/YingY0RHL18}
Z.~Ying, J.~You, C.~Morris, X.~Ren, W.~L. Hamilton, and J.~Leskovec.
\newblock Hierarchical graph representation learning with differentiable
  pooling.
\newblock In {\em Advances in Neural Information Processing Systems 31: Annual
  Conference on Neural Information Processing Systems 2018, NeurIPS 2018,
  December 3-8, 2018, Montr{\'{e}}al, Canada}, pages 4805--4815, 2018.

\bibitem{DBLP:conf/eccv/ZeilerF14}
M.~D. Zeiler and R.~Fergus.
\newblock Visualizing and understanding convolutional networks.
\newblock In D.~J. Fleet, T.~Pajdla, B.~Schiele, and T.~Tuytelaars, editors,
  {\em Computer Vision - {ECCV} 2014 - 13th European Conference, Zurich,
  Switzerland, September 6-12, 2014, Proceedings, Part {I}}, volume 8689 of
  {\em Lecture Notes in Computer Science}, pages 818--833. Springer, 2014.

\bibitem{DBLP:conf/aaai/ZhangCNC18}
M.~Zhang, Z.~Cui, M.~Neumann, and Y.~Chen.
\newblock An end-to-end deep learning architecture for graph classification.
\newblock In {\em Proceedings of the Thirty-Second {AAAI} Conference on
  Artificial Intelligence, (AAAI-18), the 30th innovative Applications of
  Artificial Intelligence (IAAI-18), and the 8th {AAAI} Symposium on
  Educational Advances in Artificial Intelligence (EAAI-18), New Orleans,
  Louisiana, USA, February 2-7, 2018}, pages 4438--4445. {AAAI} Press, 2018.

\bibitem{zhang2020backdoor}
Z.~Zhang, J.~Jia, B.~Wang, and N.~Z. Gong.
\newblock Backdoor attacks to graph neural networks, 2020.

\end{thebibliography}

%%
%% If your work has an appendix, this is the place to put it.
%\appendix

\end{document}